# Domain Adaptive Transfer Attack (DATA)-based Segmentation Networks for Building Extraction from Aerial Images

Younghwan Na, *Student Member, IEEE*, Jun Hee Kim, *Student Member, IEEE*,
Kyungsu Lee, *Student Member, IEEE*, Juhum Park, *Member, IEEE,*
Jae Youn Hwang, *Member, IEEE* and Jihwan P. Choi, *Senior Member, IEEE*

*Abstract*— **Semantic segmentation models based on convolutional neural networks (CNNs) have gained much attention in relation to remote sensing and have achieved remarkable performance for the extraction of buildings from high-resolution aerial images. However, the issue of limited generalization for unseen images remains. When there is a domain gap between the training and test datasets, CNN-based segmentation models trained by a training dataset fail to segment buildings for the test dataset. In this paper, we propose segmentation networks based on a domain adaptive transfer attack (DATA) scheme for building extraction from aerial images. The proposed system combines the domain transfer and adversarial attack concepts. Based on the DATA scheme, the distribution of the input images can be shifted to that of the target images while turning images into adversarial examples against a target network. Defending adversarial examples adapted to the target domain can overcome the performance degradation due to the domain gap and increase the robustness of the segmentation model. Cross-dataset experiments and the ablation study are conducted for the three different datasets: the Inria aerial image labeling dataset, the Massachusetts building dataset, and the WHU East Asia dataset. Compared to the performance of the segmentation network without the DATA scheme, the proposed method shows improvements in the overall IoU. Moreover, it is verified that the proposed method outperforms even when compared to feature adaptation (FA) and output space adaptation (OSA).**

## I. INTRODUCTION

WITH the development of sensor technology, many remote sensing images have become available [1]-[3], and such images contain highly detailed information, including the shapes of buildings, vehicles, and roads among other elements. To utilize aerial images, extracting the objects of interest from them is necessary. However, extracting objects from images obtained under different conditions (e.g., locations, times, weather) can be difficult, even with identical objects. Especially for buildings, with large variations in the sizes, colors, and angles, misclassifications can arise [4]. Thus, building extraction from aerial images is mainly performed by a hand-digitizing process [5], but this is laborious and time-consuming.

Currently, convolutional neural networks (CNNs) [6], replacing hand-crafted feature extraction with learnable filters, show remarkable performance in a wide range of areas, such as image classification [7]-[12] and object detection [13]-[16]. However, CNNs face structural limitations when used for the pixel-wise labeling of images. During the feature extraction process, using consecutive pooling layers causes a loss of spatial information (e.g., the boundaries or edges of objects). Thus, most CNN architectures are good at recognizing objects but rarely succeed in localizing objects precisely [17]. In the pixel-wise labeling of aerial or satellite images, these problems become more challenging. Misclassifications of object boundaries hamper the overall performance of semantic segmentation systems. To achieve both recognition and localization, several types of architecture for aerial images have been studied [17]-[23]. A refinement method to rectify the coarse output of CNNs has been proposed [17]. Fusing multi-level feature maps to achieve accurate boundary inference outcomes is also discussed [18]. On the other hand, in other work [19], adversarial networks [20] are utilized to generate label maps that cannot be distinguished from ground truth. In another study [21], a Sobel detector [22] and a fully convolutional network are jointly used to rectify the semantic contour. Additionally, combining a pyramid pooling layer [24] with the U-Net architecture [25] is proposed [23].

Although these methods improve the performance of deep-learning models, the trained models fail to segment objects for unseen images when there is a domain gap between the training and test data. To solve this problem, a domain adaptation method is proposed [26], [27]. Numerous domain adaptation methods have been developed based on CNN classifiers to address the domain shift problem between the source and target domain. In one such study [26], convolutional neural networks are trained by gradient reversal to adapt to the target domain. The goal of the trained model is to ensure that the results of feature extraction are such that a discriminator cannot classify between the training and test domains. Attempts [28], [29] have been made to apply this feature adaptation (FA) in the field of image segmentation. In other work [30], an output space adaptation (OSA) approach utilizing structured outputs that contain spatial similarities between the source and target domains is proposed, showing better semantic segmentation performance than the FA method. However, it is difficult to train adversarial networks used in these methods, and the risk of model collapse is inherent [19], [31].

Machine-learning models are vulnerable to an adversarial example in the form of a perturbed input designed to mislead a model [32], [33]. Augmenting training data with adversarial examples increases the degree of robustness [34]. In one approach [35], a feed-forward neural network generates adversarial examples against a targeted network. The model restricts the adversarial example such that it is perceptually similar to the original image and alters the prediction of the targeted network of the resulting image. A study has been proposed to enhance the generalization of the deep learning model in the unseen domain by inserting adversarial examples that are difficult for the current model into the dataset [36].

In this paper, we suggest the domain adaptive transfer attack (DATA)-based segmentation networks for building extraction from high-resolution aerial images. The proposed training method uses domain adaptation-based adversarial training for aerial image segmentation. Because both domain transfer and adversarial attack concepts are used jointly, it is possible to force the distribution of the generated adversarial examples toward that of the target domain. Transformed images are likely to cause a semantic segmentation network to make a misprediction, whereas they are close to the target distributions. Augmenting the training data with the DATA scheme can improve the generalization of the segmentation model and mitigate the domain gap problem between the source and the target domains.

The main contributions of this paper are summarized below.

1) The DATA scheme utilizes both domain transfer and adversarial attack concepts to force the direction of image generation. This neural network-based attack not only tricks the target segmentation model but also uncovers the mapping from representations of the source domain to those of the target. Based on the DATA scheme, the training dataset is adapted to the test dataset, as proved by the t-stochastic neighbor embedding (t-SNE) scheme [37]. The converted images generated by the DATA scheme are used to train the base segmentation network developed by us.

2) Cross-dataset experiments and the ablation study are

conducted. The segmentation model trained by the DATA scheme shows higher mean intersection over union (mIoU) scores compared to the outcomes without the DATA scheme, achieving greater improvements than both FA and OSA schemes.

The rest of this paper is organized as follows. In Section II, semantic segmentation models for building extraction from high-resolution aerial images are studied to address the relationship between the domain gap and the segmentation performance. In Section III, the proposed augmentation system is discussed. In Section IV, training results with/without the DATA scheme are described. Finally, we conclude our paper in Section V.

## II. BASIC ARCHITECTURE FOR SEMANTIC SEGMENTATION

### A. Semantic Segmentation Systems

The goal of semantic segmentation is to assign each pixel a semantic label (e.g., buildings and background) in images. One of the basic architectures used to segment images is the fully convolutional network (FCN) [38], which is a modified version of a CNN. Due to the excellent performance of FCNs, recently designed segmentation models are mostly based on the FCN architecture. The FCN consists of an encoder for feature extraction and a decoder that up-samples the extracted features to the original image size. In recently proposed models, more efficient encoder models are adopted to improve the performance. Moreover, integrating the extracted features with certain independent modules, such as a pyramid pooling layer [24] or a summation-based skip connection [39], has been utilized. We also suggest a semantic segmentation type of architecture, known as Residual DenseNet with Squeeze-and-Excitation (SE) blocks [40] (RD-SE). This network has an efficient encoder architecture and an integrating feature module. Fig. 1 illustrates the overall architecture for building extraction. Residual dense blocks [41] for fusing multiple feature maps to compensate for the spatial loss which arises during the process of feature extraction are adopted. They allow gradients to flow smoothly across multiple layers. Skip connections are densely connected to strengthen feature propagation [42], Also, to utilize the interdependencies between the channels of convolutional features, SE blocks are used at the end of residual dense block, as shown in Fig. 1 (c). Channel-wise feature responses are adaptively recalibrated. The designed model is trained by the Inria aerial image labeling dataset [1] and compared with other deep learning models.

### B. Inria Aerial Image Labeling Dataset

The Inria aerial image labeling dataset [1] consists of 360 orthorectified aerial images for an area of 810 $km^2$. There are 5,000 x 5,000 pixels in each image, and the resolution is 0.3 m. Ground truth data has two semantic classes (buildings and background). The dataset covers ten different cities and incorporates various urban landscapes and settlements. It is

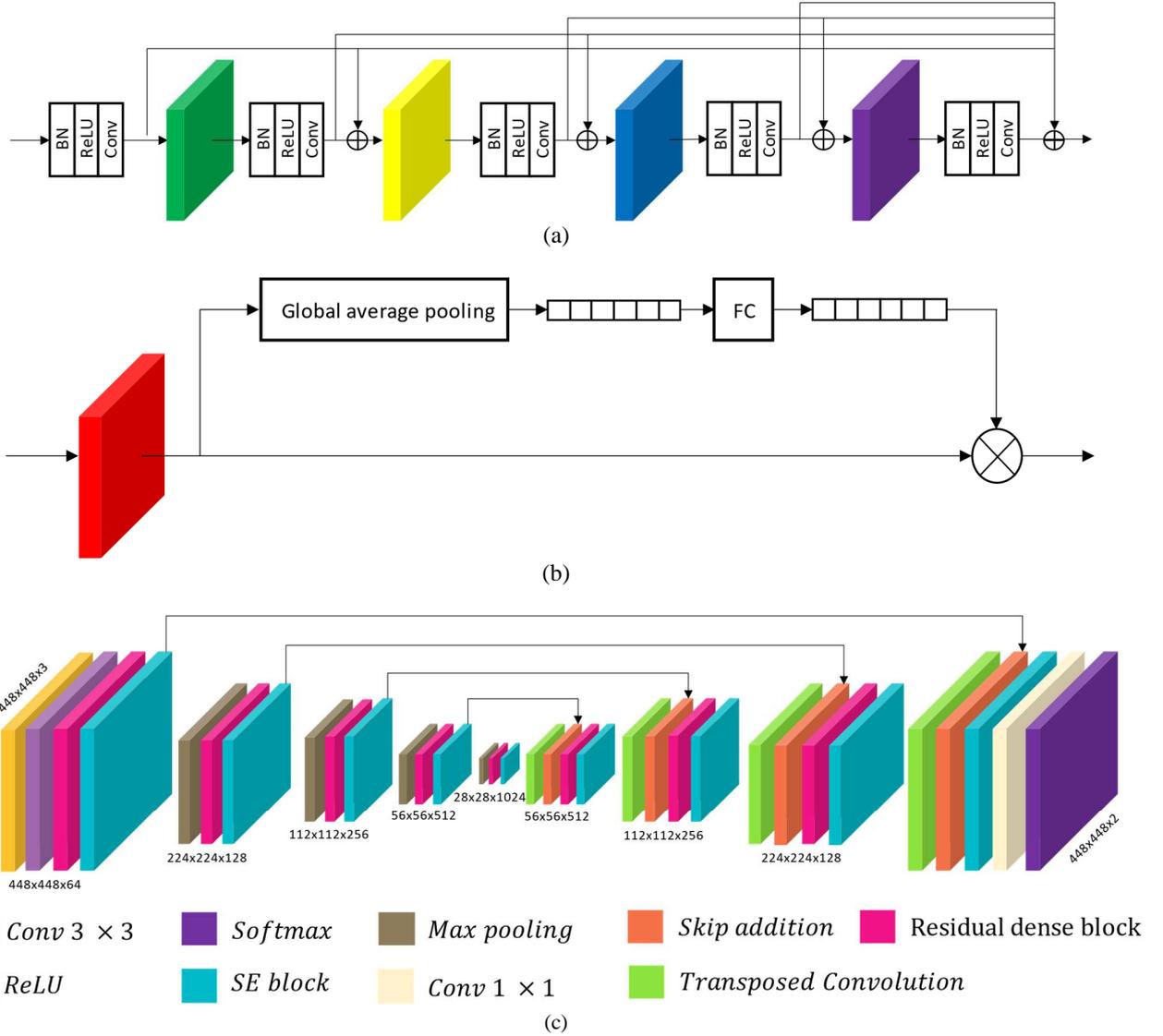

*BN: Batch normalization [43], Conv: Convolution, FC: Fully connected layer, ReLU: Rectified linear unit

Fig. 1. Overall architecture for semantic segmentation: (a) Residual dense block (b) Squeeze and excitation block, and (c) System architecture for building extraction

divided into 155 images for training, 25 images for validation and the remaining 180 images for testing. The regions used in the training set are Austin, Chicago, Kitsap County, Western Tyrol and Vienna, and the regions in the testing set are Bellingham, Bloomington, Innsbruck, San Francisco and Eastern Tyrol.

### C. Training Setup

Our model is implemented based on TensorFlow [44]. We use the Adam optimizer [45] with a learning rate of $\alpha = 1e\text{-}4$ and the exponential decay rates for the moments of estimates of $\beta_1 = 0.9$ and $\beta_2 = 0.999$. The epsilon and momentum values of the batch normalization layer are 0.99 and 0.001, respectively. As the activation function for hidden layers, the rectified linear unit (ReLU) is used. The segmentation network is trained with binary cross-entropy loss and modified IoU loss between the prediction results and the ground truth:

$$\mathcal{L}_S(P_s, Y) = \mathcal{L}_{BCE}(P_s, Y) + \mathcal{L}_{IoU_{new}}(P_s, Y), \qquad (1)$$

where $\mathcal{L}_{BCE}(P_s, Y)$ is the binary cross-entropy loss, and $\mathcal{L}_{IoU_{new}}(P_s, Y)$ is the modified IoU loss, with segmentation prediction $P_s$ and ground truth $Y$. The binary cross-entropy loss $\mathcal{L}_{BCE}(P_s, Y)$ can be written as

$$\mathcal{L}_{BCE}(P_s, Y) = -\sum_{h,w}\sum_{c\in C} Y^{(h,w,c)} \cdot \log\big(P_s^{(h,w,c)}\big). \qquad (2)$$

We modified the IoU loss [46] for semantic segmentation. If the number of object pixels in a batch is low, a misclassification of objects by a few pixels causes a large IoU loss. Thus, the conventional IoU loss is multiplied by the ratio of the union area, as given by

$$\mathcal{L}_{IoU_{new}}(P_s, Y)$$

$$= \mathcal{L}_{IoU_{old}}(P_s, Y) \cdot \frac{\sum(P_s + Y - P_s \cdot Y)}{N}$$

$$= \left(1 - \frac{\sum(P_s \cdot Y)}{\sum(P_s + Y - P_s \cdot Y)}\right) \cdot \frac{\sum(P_s + Y - P_s \cdot Y)}{N}$$

$$= \frac{\sum(P_s + Y - P_s \cdot Y) - \sum(P_s \cdot Y)}{N}. \qquad (3)$$

TABLE I
Test Results on Inria Aerial Image Labelling Test set

| Method | Bellingham | Bloomington | Innsbruck | San Francisco | East Tyrol | Overall IoU |
|---|---|---|---|---|---|---|
| Building-A-Net [19] | 65.50 | 66.63 | 72.59 | 76.14 | 71.86 | 72.36 |
| Dual-resolution U-Nets [47] | 70.74 | 66.06 | 73.17 | 73.57 | 76.06 | 72.45 |
| UNetPPL [23] | 69.39 | 66.90 | 71.45 | 77.07 | 73.08 | 73.27 |
| Sobel Heuristic Kernel [21] | 70.73 | 69.98 | 76.74 | 76.73 | 79.09 | 75.33 |
| Koki Takahashi [48] | 74.15 | 75.55 | 78.62 | 80.65 | 80.80 | 78.80 |
| RD-SE (Ours) | 73.68 | 78.91 | 78.80 | 80.90 | 81.23 | 79.35 |
| ICT-Net [49] | **74.63** | **80.80** | **79.50** | **81.85** | **81.71** | **80.32** |

$\mathcal{L}_{IoU_{old}}(P_s, Y)$ is the conventional IoU loss and $N$ is the number of pixels. The network takes in patches of $448 \times 448$, with the patches selected randomly. In this case, 155 images are used for training and 180 images are used for testing. We set the batch size to 4. Data augmentation methods, specifically rotation, flip, brightness changes, blur, and Gaussian noise, are used. The experiments are run on an Intel Core 6 i7-7820X CPU at 3.6 GHz with two NVIDIA GeForce TITAN Xp GPUs (12 GB).

### D. Test Results and Comparison with other Architectures

The test results for the Inria test dataset are summarized in Table I. The designed model achieves an overall IoU score of 79.35. We compare other architectures trained by the Inria labeling dataset and prove that the performance of the designed model is comparable to those of state-of-the-art models. Although the performance of designed model is comparable with state-of-the-art models, it has limited generalization power for cities are not included in the training images, as shown in the following subsection.

### E. Test Results with other Datasets

One of the main problems in semantic segmentation is the limited generalization, referring to the failure to segment with unseen image domains. This is caused by the domain gap between the training data (source) and the test data (target). For example, the appearance for buildings in aerial images in different cities can vary. Thus, a model trained with one city can misclassify buildings in other cities. To show the relationship between the domain gap and the segmentation performance for different datasets, we test two additional datasets, the Massachusetts building [2] dataset and the WHU [3] dataset, which are not parts of the Inria training dataset.

The Massachusetts building dataset [2] consists of aerial images of the Boston area. There are 151 images in total and the spatial resolution is 1.0-m. Each image has 1,500 x 1,500 pixels for an area of 2.25 $km^2$. The dataset is split into a training set of 137 images, a test set of ten images and a validation set of four images. Here, we used 104 training images that do not contain any white background.

The WHU dataset [3] is composed of six neighboring satellite images. It covers an area of 550 $km^2$ in East Asia with 2.7 m spatial resolution. Each image has 512 x 512 pixels. The vector building map is fully delineated manually in the ArcGIS software. The dataset contains a total of 29,085 buildings. We only used tiles that had a building provided by the organizer. The training images are 3,135 tiles and the number of tiles used for test is 903.

We visualize the data distribution for each dataset using the t-SNE approach [37]. Fig. 2 illustrates the data distribution for the Inria, Massachusetts, and WHU datasets. Each dataset forms a distinct cluster. We conduct cross-dataset experiments to examine the relationship between the domain gap and the performance of the semantic segmentation model. Three neural networks are trained on each of the train sets. Then, we evaluate each on the Inria validation set, the Massachusetts test set and the WHU test set. The Inria validation dataset, which has the same cities as the training dataset does, was used for test because the cities constituting the Inria test dataset and those of the training dataset are different. The test results are summarized in Table II. From the test results, it is verified that the performance of building segmentation deteriorates if there

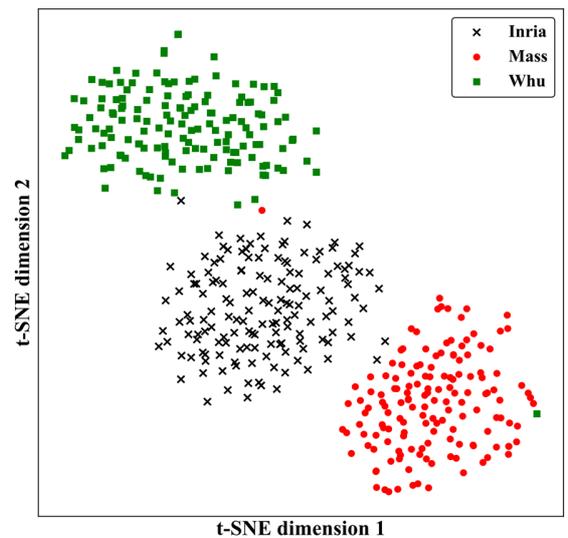

*Mass: Massachusetts

Fig. 2. Visualization for each dataset

TABLE II
TEST RESULTS ON EACH DATASET

| Test set / Train set | Inria | Mass | WHU |
|---|---|---|---|
| Inria | 80.77 | 48.46 | 12.87 |
| Mass | 40.36 | 73.95 | 18.51 |
| WHU | 42.33 | 7.22 | 71.42 |

is a domain gap between the training and test data. To reduce the domain gap between the training and test data, we propose the domain-adaptive transfer attack (DATA) scheme, which is discussed in Section III.

### F. Related Work for Domain Adaptation

There are several ways to handle the domain adaptation problem. For image classification, the domain-adversarial neural net (DANN) [26] has been newly proposed to extract domain-invariant features. It aligns the feature distribution between the source and target images. Similar approaches have been studied using adversarial learning in the feature space for image semantic segmentation [28], [29]. Unlike in the classification problem, FA for semantic segmentation has one challenge related to the complexity of high-dimensional features, which are needed to encode diverse visual cues, including the appearance, shape and context. Because the output space contains rich information spatially and locally, methods for adapting pixel-level prediction tasks rather than using the feature space have been developed. [30]. Addressing the pixel-level domain adaptation problem in the output space surpasses FA. These domain adaptation methods are based on training adversarial networks. GAN-based domain adaptation methods have a high risk of model collapse due to imbalance between adversarial networks. [19], [31]. The characteristics of each aerial image dataset are obvious, making it easy for the discriminator to overwhelm the generator. This causes the gradient of the generator to vanish or become very large. Therefore, we propose DATA-based segmentation networks for building extraction from aerial images. The next section covers the overall system of the DATA scheme.

### III. DOMAIN ADAPTIVE TRANSFER ATTACK (DATA)

In the previous section, we discussed the relationship between the domain gap and the performance of the segmentation model. Due to the domain gap, deep learning models fail to segment buildings from datasets that are not included in the training data. To solve this problem, we propose segmentation networks based on the domain adaptive transfer attack (DATA) scheme in this section.

### A. Overview of The Proposed Model

The general idea of our DATA scheme is to train the attack model to obtain domain-adapted adversarial examples that are used for training of the segmentation model. The proposed system consists of three modules: an adversarial attack network $G$, a discriminator $D$, and a segmentation network $S$. There are two goals when training an adversarial attack network. The first is to transform source domain images to appear as if drawn from the target domain. The second is to manipulate the generated images to cause the segmentation model $S$ to make a misclassification. Adversarial examples that meet the second goal can help to increase the test-set generalization and adversarial robustness of the model [36]. To fulfil these tasks, we use jointly the domain transfer and adversarial attack concepts. Because only a small amount of change is added to the original image, stable learning is possible.

The overall system architectures are plotted in Fig. 3. The sets of source and target images are denoted as $\{\mathcal{I}_S\}$ and $\{\mathcal{I}_T\} \in \mathbb{R}^{H \times W \times 3}$, where $H$ is the height of each image and $W$ is the width of each image. The source image $I_s$ is forwarded as the input of the adversarial attack network $G$. The overall architecture of the adversarial attack model $G$ is U-Net, which is commonly used as a generator, except that we add a skip connection to the last layer. This skip connection maintains a semantic space using the summation of the input and perturbation. Note that $\lambda$ is the weighted value for the combination of the input and perturbation. The converted image $G(I_s; \theta_G)$ is then used as the input of segmentation model $S$ and discriminator $D$, which distinguishes between the source and target images, as explained in Subsection III-C. Because the second goal of the adversarial attack model $G$ is to cause the segmentation model $S$ to make a mistake, the segmentation prediction $P_s \in \mathbb{R}^{H \times W \times 2}$ should be far from the ground truth $Y$. For a domain transfer, $G$ should cause the discriminator $D$ to classify the input as a target image.

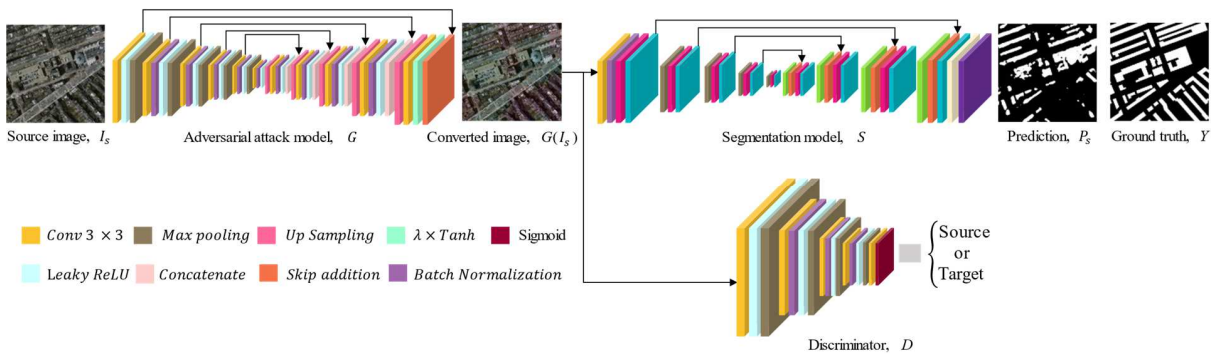

Fig. 3. Overall architecture of the domain-adaptive transfer attack (DATA)-based segmentation network scheme.

## B. Objective Function for the Generator

The objective of the adversarial attack network $\mathcal{L}_G$ is composed of multiple different optimization targets:

$$\mathcal{L}_G(I_s, Y) = \alpha \mathcal{L}_{inv}(I_s) + \beta \mathcal{L}_{atk}(I_s, Y) + \gamma \, \mathcal{L}_{tr}(I_s), \quad (4)$$

where $\mathcal{L}_{inv}(I_s)$ is the invariance loss, $\mathcal{L}_{atk}(I_s, Y)$ is the attack loss and $\mathcal{L}_{tr}(I_s)$ is the domain transfer loss. The goal of the adversarial attack model $G$ is to determine the value of $\theta_G^*$ that minimizes a combination of different targets. To ensure perceptual similarity with the original image, we utilize the $L_1$ norm as the invariance loss, as follows:

$$\mathcal{L}_{inv}(I_s) = \|I_s - G(I_s; \theta_G)\|_1. \quad (5)$$

As the attack loss $\mathcal{L}_{atk}$ causes the segmentation model $S$ to make a mistake, the segmentation prediction $S(G(I_s; \theta_G); \theta_S)$ corresponding to $P_s$, where $\theta_S$ is a parameter of the semantic segmentation model $S$, should be far from the ground truth $Y$. Thus, we model the adversarial attack function as follows,

$$\mathcal{L}_{atk}(I_s, Y) = -\mathcal{L}_{IoU_{new}}(P_s, Y). \quad (6)$$

$\mathcal{L}_{tr}(D(G(I_s; \theta_G); \theta_D))$ is the domain transfer loss that adapts the distribution of the converted images to the target distribution. The loss function for adapting to the target domain can be written as follows:

$$\mathcal{L}_{tr}(I_s) = \|D(G(I_s; \theta_G); \theta_D) - 1\|_2^2. \quad (7)$$

To balance attack and transfer while maintaining the semantic space, we arbitrarily set the weighted values of $\alpha, \beta$ and $\gamma$ to 2.0, 0.5 and 1.0, respectively, according to the order of importance of each objective function.

## C. Objective Function for Discriminator

The goal of the discriminator $D$ is to distinguish whether the input is from the source or target domain, as shown in Fig. 4. The objective function for the discriminator $D$ can be formulated as follows,

$$\mathcal{L}_D(I) = \mathbb{E}_{I \sim \mathcal{I}_S}[\|D(I; \theta_D)\|_2^2] + \mathbb{E}_{I \sim \mathcal{I}_T}[\|D(I; \theta_D) - 1\|_2^2], \quad (8)$$

where a parameter of the discriminator $D$, $\theta_D$, is updated to minimize $\mathcal{L}_D$ during the training phase. The discriminator $D$ does not compete with $G$, but merely provides information about the target domain to the adversarial attack model $G$.

## D. Training the Adversarial Attack Model & Discriminator

Because holding the semantic space for the original images is crucial during data augmentation for segmentation, we set the weighted value $\lambda$ to $\frac{1}{15}$, during the training process. Experiments on the performance impact of changing this value are also discussed in the next section. The trained $G$ converts training images to adversarial examples without losing the semantic space. Fig. 5 shows sample images of the output of $G$. The perturbed images trick the target segmentation network. In other words, these data are fairly difficult examples for existing models, and will help to expand the generalization capability of the segmentation model.

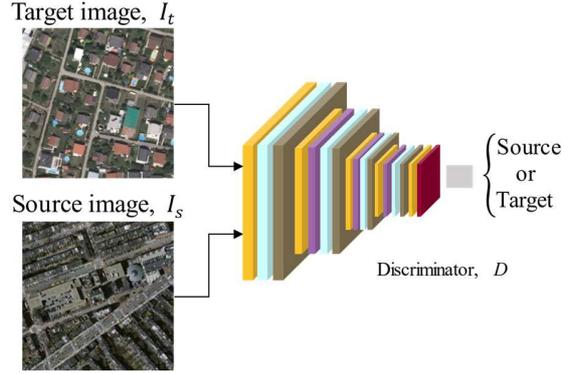

Fig. 4. Overall architecture of the discriminator, $D$

The adversarial attack model $G$ is learned so that the $D$ predicts that the converted image comes from the target domain. At the same time, the D trains to classify whether the image is from the source or target domain. The discriminator learns the train image of the dataset that is the objective of adaptation as the target image. Ground-truths of target datasets are not used at the training time. The effectiveness of the DATA scheme can be verified by the t-SNE approach. Data distributions for each dataset are plotted in Fig. 6. Except for the case between Massachusetts and WHU with very large feature differences, other domain transfers show sufficiently successful results. The distribution of converted images is shifted to target domain, indicating that the domain gap is reduced by the DATA scheme. Then, we utilize the converted images to expand the generalization power of the semantic segmentation network $S$; i.e., both the training data and the converted data are used in the training procedure. DATA-based adversary training is discussed in the following section in detail.

## IV. DATA-BASED ADVERSARIAL TRAINING & RESULTS

From the results presented in the previous section, the DATA scheme can replace the image from the source domain with an adversarial example in the target domain. In this section, to expand the generalization of the segmentation network and reduce the domain gap, we utilize the adversary training method. These results are discussed below. To verify effectiveness of the DATA scheme, we apply FA and OSA to a segmentation model separately and compare the results. Ablation studies on the different components of the proposed method are conducted, e.g., domain transfer-based augmentation and adversarial attack-based augmentation. This section also contains extended experiments in various environments.

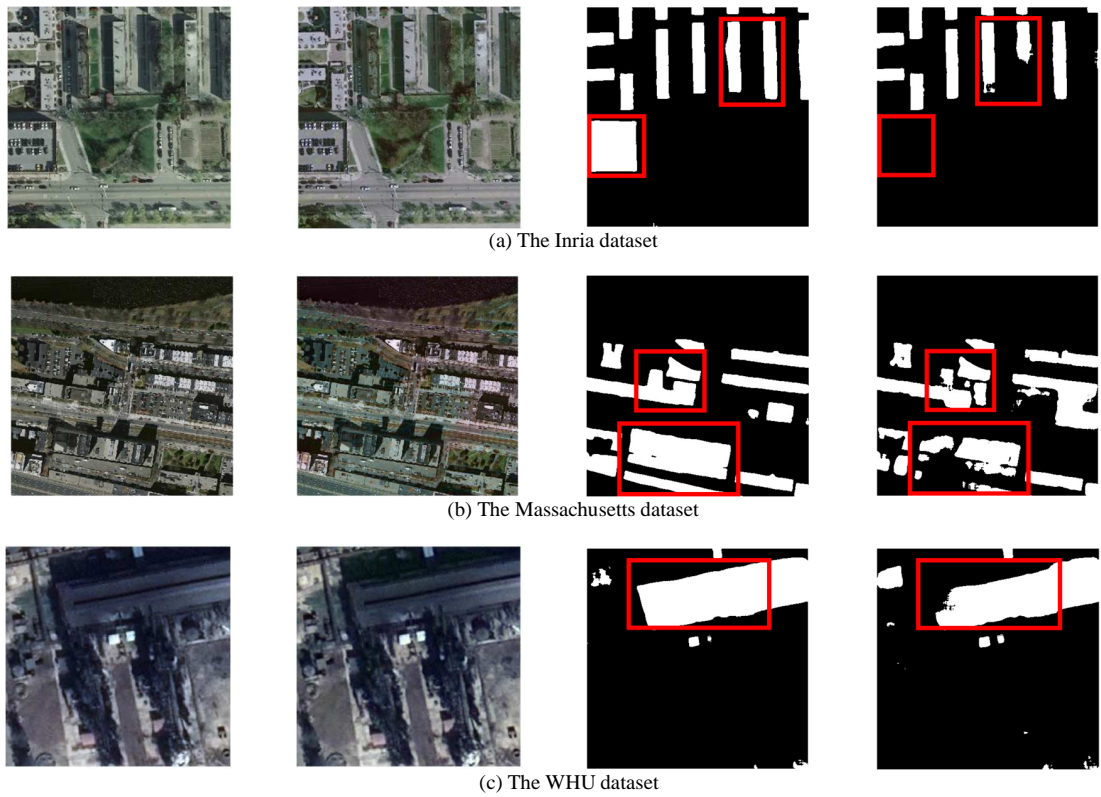

(a) The Inria dataset

(b) The Massachusetts dataset

(c) The WHU dataset

Fig. 5. Sample images and predictions with/without adversarial attack model **G**. The first column is the input of **G** and the second column is the output of **G**. The third and fourth columns are the predictions of **S** using the first and second column images, respectively.

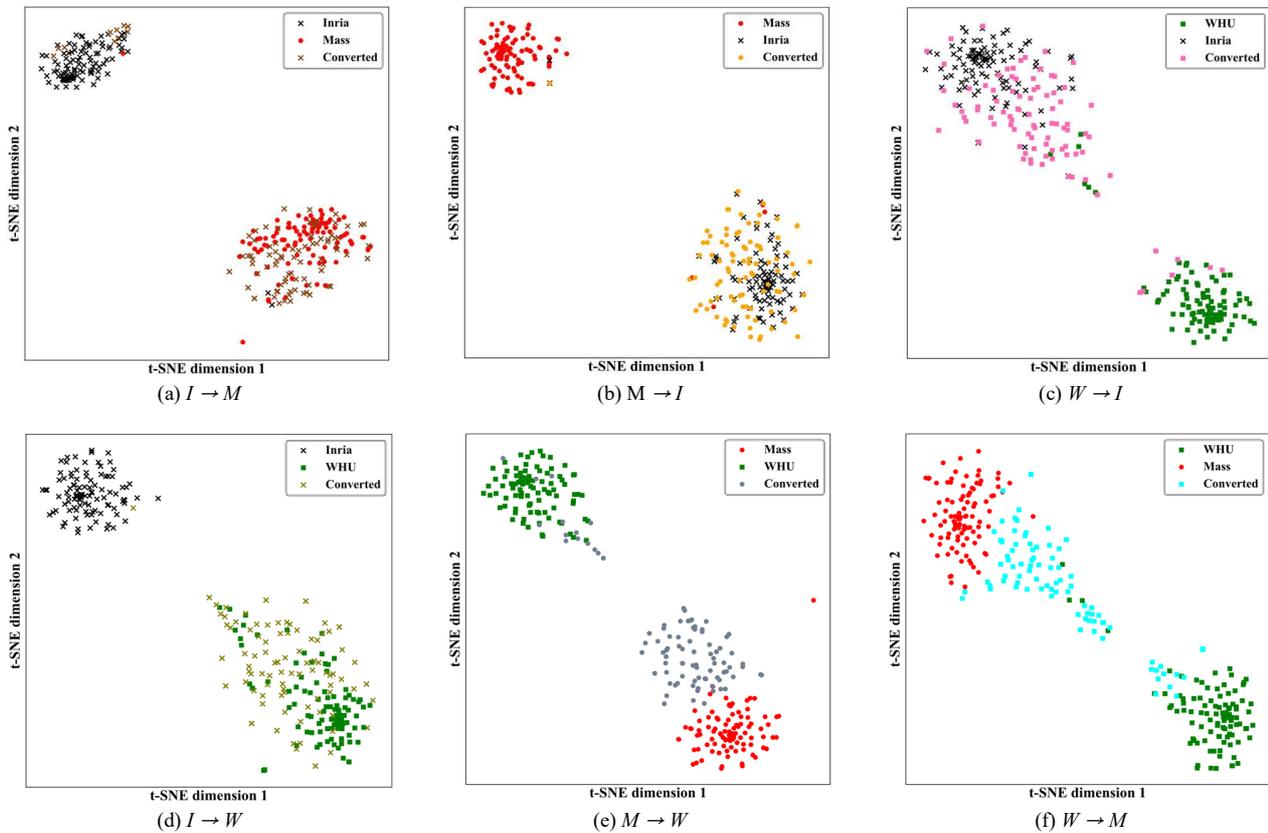

(a) $I \rightarrow M$

(b) $M \rightarrow I$

(c) $W \rightarrow I$

(d) $I \rightarrow W$

(e) $M \rightarrow W$

(f) $W \rightarrow M$

Fig. 6. Visualization for each dataset with converted images: The first legend of each graph is the source domain dataset, the second is the target domain dataset, and the third is the converted source image dataset that is output of attack model **G**. The marker shape of the converted dataset is same as the marker of the source domain. *I, M and W* stand for Inria, Massachusetts and WHU dataset.

TABLE III
IOU SCORES FOR EACH ADAPTATION METHOD

| Source Dataset | Target Dataset | No Adaptation | FA | OSA | Attack | Transfer | DATA(Ours) |
|---|---|---|---|---|---|---|---|
| Inria | Mass | 48.46 | 43.27 | 52.90 | 52.31 | 52.05 | **56.29** |
|  | WHU | 12.87 | 12.69 | 3.39 | 16.82 | 21.92 | **26.65** |
| Mass | Inria | 40.36 | 45.69 | 45.56 | 45.60 | 47.56 | **51.29** |
|  | WHU | 18.51 | 21.30 | 12.17 | 18.78 | **24.23** | 23.10 |
| WHU | Mass | 7.22 | 13.24 | 17.42 | **19.61** | 10.86 | 14.52 |
|  | Inria | 42.33 | 38.27 | 36.77 | 45.92 | **49.66** | 47.79 |

### A. Adversary Training Setup

The segmentation model $S$ is trained with both the source dataset and the converted images by the proposed DATA scheme. For fast learning, we freeze the encoder and fine-tune the segmentation model $S$, as discussed in Section III. We introduce $k$ as a switching parameter that determines whether to activate the adversarial attack model $G$. The adversarial attack model $G$ operates for every $k$-iteration and the converted images are used to train the segmentation model $S$. The reason for introducing this parameter is to prevent overfitting on adversarial examples. Note that we set the value of $k$ to 1 for this experiment. The adversary training process is summarized in **Algorithm 1**.

---
**Algorithm 1** Adversary Training Process
---
**Input:** source dataset and learned weights $\theta_S$
**Output:** learned weights $\theta_S$

01: **for** $t = 1, \ldots, $ T **do**
02:    Sample $I_s$ from dataset $\{\mathcal{I}_S\}$ and ground truth $Y$
03:    **Compute** $\mathcal{L}_D$
04:    **Update** the weights of discriminator $\theta_D$
          with respect to $\mathcal{L}_D$
05:    Sample $I_T$ from dataset $\{\mathcal{I}_T\}$
06:    **Compute** $\mathcal{L}_D$
07:    **Update** the weights of discriminator $\theta_D$
          with respect to $\mathcal{L}_D$
08:    **Compute** $\mathcal{L}_G$
09:    **Update** the weights of attack model $\theta_G$
          with respect to $\mathcal{L}_G$
10: **end for**
11: **for** $j = 1, \ldots, $ J **do**
12:    Sample $I_s$ from dataset $\{\mathcal{I}_S\}$ and ground truth $Y$
13:    Generate the prediction label map $P_s$ as $S(I_s; \theta_S)$
14:    **Compute** $\mathcal{L}_S$
15:    **Update** the weights of segmentation model $\theta_S$ with
          respect to $\mathcal{L}_S$
16:    **if** $j$ % $k == 0$ **do**
17:       Operate the adversary attack model $G$ and convert
             input image $G(I_s; \theta_G)$
18:       Generate the prediction label map $P_s$
             as $S(G(I_s; \theta_G); \theta_S)$,
19:       **Compute** $\mathcal{L}_S$
20:       **Update** the weights of segmentation model $\theta_S$
             with respect to $\mathcal{L}_S$
21:    **end if**
22: **end for**
---

### B. Comparison with Other Methods

The test results for the segmentation network $S$ trained using the source dataset and the converted images are summarized in Table III. Source and target represent the datasets used to train and test the segmentation model, respectively. By applying FA and OSA, some of dataset experiments can perform well thanks to domain adaptation, whereas others cannot mainly due to model collapse. Learning is not stable because it is difficult to balance a generator and a discriminator. As can be seen in the ablation study, the average IoU improvement of attack, transfer and DATA methods are 4.88, 6.09 and 8.32, respectively, compared to the values of "No Adaptation", which are the IoU scores of the base model without the adaptation method. Although our DATA scheme does not show the highest IoU for every source-target pair, it achieves the best average performance gain. Because the converted images for each dataset are relatively close to the target distribution as compared to the training data, it leads to improve model test-set generalization.

We also compare the resource allocation of our method with other algorithms. Table IV shows the memory and time consumption. The second column lists the maximum memory required to run the deep learning algorithm at batch size 1. The third column shows the running time for an iteration. In the proposed model, since the part for training an attacker and a discriminator (left) and the part for training a semantic segmentation model (right) are separated, the time measurement is indicated separately. The algorithm without domain adaptation uses the least resources. The GAN-based domain adaptation increases memory usage significantly because all the models are trained in at one time. Since the proposed method trains modules separately, the memory increase is small, but the time requirement is quite notably increased.

In Fig. 7 to Fig. 12, the prediction results of all dataset experiments are illustrated for the FA, OSA, Attack, Transfer and DATA scheme. The model trained by the DATA scheme properly assigns building pixels with semantic labels.

TABLE IV
RESOURCE ALLOCATION COMPARISON

| Method | Memory (MiB) | Time (Sec / Iteration) | |
|---|---|---|---|
| RD-SE | 7,569 | 0.48 | |
| FA | 11,151 | 0.64 | |
| OSA | 11,149 | 0.55 | |
| Attack | 7.569 | 0.33 | 048 |
| Transfer | 8,077 | 0.14 | 0.49 |
| DATA | 8,077 | 0.39 | 0.49 |

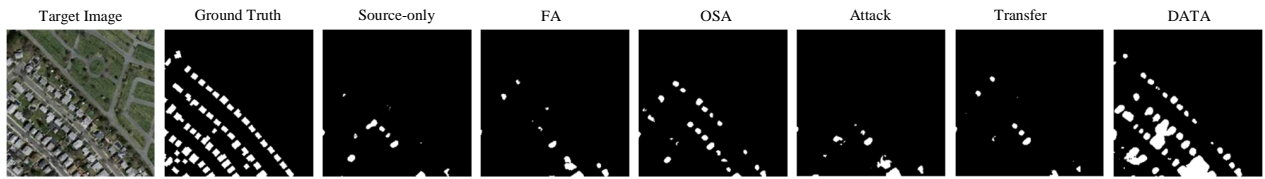

Fig. 7.  Example results of adapted segmentation for Inria to Massachusetts: "source-only" corresponds to training on the Inria dataset only.

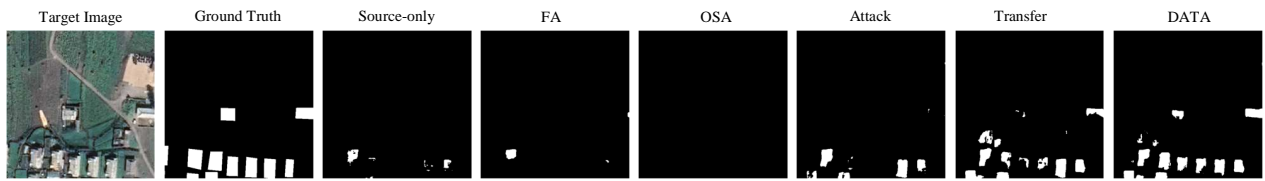

Fig. 8.  Example results of adapted segmentation for Inria to WHU: "source-only" corresponds to training on the Inria dataset only.

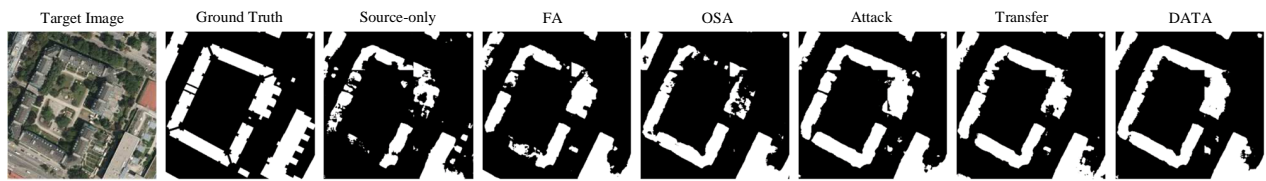

Fig. 9.  Example results of adapted segmentation for Massachusetts to Inria: "source-only" corresponds to training on the Massachusetts dataset only.

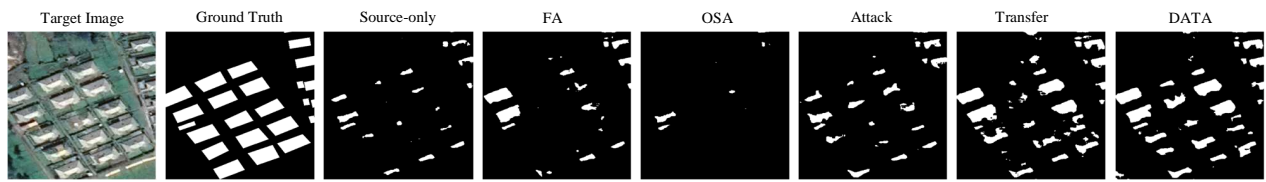

Fig. 10.  Example results of adapted segmentation for Massachusetts to WHU: "source-only" corresponds to training on the Massachusetts dataset only.

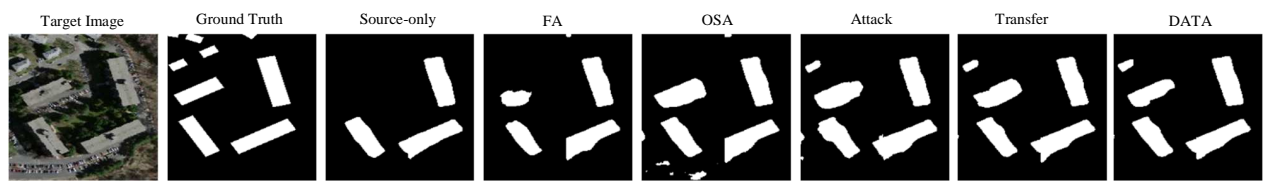

Fig. 11.  Example results of adapted segmentation for WHU to Massachusetts: "source-only" corresponds to training on the WHU dataset only.

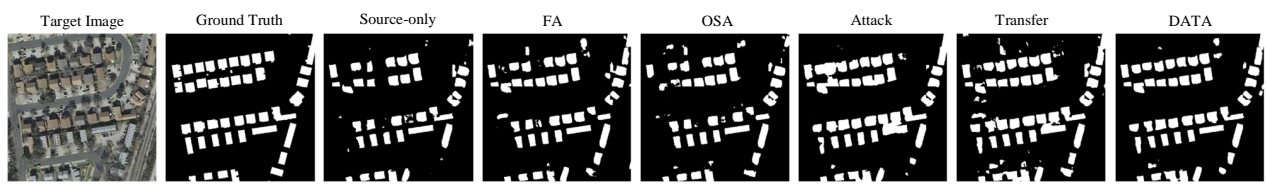

Fig. 12.  Example results of adapted segmentation for WHU to Inria: "source-only" corresponds to training on the WHU dataset only.

TABLE V
TEST RESULTS FOR CHANGING $\lambda$

| Source Dataset | | Inria | | Mass | | WHU | |
|---|---|---|---|---|---|---|---|
| Target Dataset | | Mass | WHU | Inria | WHU | Mass | Inira |
| No Adaptation | | 48.46 | 12.87 | 40.36 | 18.51 | 7.22 | 42.33 |
| $\lambda$ | 1 / 15 | 56.29 | 26.65 | **51.29** | 23.10 | 14.52 | 47.79 |
| | 1 / 12 | 54.43 | 32.32 | 51.02 | 24.93 | **17.57** | 43.67 |
| | 1 / 10 | **57.20** | 32.30 | 45.36 | **31.25** | 12.77 | **46.87** |
| | 1 / 8 | 53.86 | **38.42** | 46.70 | 25.99 | 17.11 | 44.73 |

*C. Extended Experiments in Various Environments*

Additional experiments in various environments are conducted to verify the effectiveness of the DATA scheme. The first one compares the impact of changing $\lambda$, the ratio of perturbation. Table V shows IoU scores according to the change of $\lambda$ in the original image. The higher the perturbation ratio is, the easier domain transfers and attacks become, but the more damages the original image has. On the other hand, if $\lambda$ is small, it is easy to maintain the semantic space of the original image, but attacks and transfers do not work well. The optimal $\lambda$ for each dataset experiment can be determined experimentally. All the cases show performance improvements compared to those without the DATA scheme. By tuning the parameters, various datasets can be obtained.

We also apply the DATA sceme to another semantic segmentation model to show the versatility of the proposed method. The segmentation model adopted in this experiment is the ICT-Net, a state-of-the-art model from the Inria labeling dataset. The FC-Densenet 45, which has fewer layers than the original model due to the memory allocation problem, is used as the backbone network for the ICT-Net. Similarly, cross-dataset experiments were conducted on each of the three datasets. Experimental results are summarized in Table VI. As there are overall performance improvements in all the cases, it is verified that the DATA method can be applied to other image segmentation models effectively.

TABLE VI
TEST RESULTS FOR ICT-NET

| Source Dataset | Inria | | Mass | | WHU | |
|---|---|---|---|---|---|---|
| Target Dataset | Mass | WHU | Inria | WHU | Mass | Inira |
| No Adaptation | 39.44 | 14.75 | 25.67 | 8.05 | 8.22 | 30.93 |
| DATA | 45.36 | 27.87 | 38.14 | 19.28 | 9.50 | 34.79 |

V. CONCLUSION

In this paper, we investigated the relationship between the domain gap and the semantic segmentation outcome. The designed deep learning model can extract buildings from high-resolution images with performance levels comparable to state-of-the-art methods. For other datasets, however, it fails to segment buildings due to the domain gap between the training and test datasets. We visualized data distributions for each dataset using the t-SNE method. Three datasets (i.e., the Inria aerial image labeling dataset, the Massachusetts building dataset, and the WHU dataset) embedded in the t-SNE space clearly have domain gaps. Thus, the performance of the segmentation model trained by one dataset deteriorates with other datasets. To reduce the domain gap between the training and test data and to expand the generalization of the model, we proposed segmentation networks based on the domain-adaptive transfer attack (DATA) scheme with the combined use of both the domain transfer and adversarial attack concepts. Domain-adapted adversarial examples are generated by the DATA scheme. The target segmentation model adapts the domain gap and shows increased performance by training with transformed images. Experiment results show that the generalization power for unseen dataset is enhanced by the effectiveness of the segmentation model based on the DATA scheme. Moreover, with the proposed scheme, we compared performance outcomes with FA and OSA and conducted ablation study. RD-SE with the DATA scheme improved the overall IoU of buildings for unseen images. On the other hand, RD-SE with FA or OSA led to deteriorated performances on building extraction tasks with some datasets due to the difficulty of training GAN. In aerial images, each dataset has significant feature differences, which increases the risk of model collapse. We attempt to transfer domains using an adversarial attack concept that changes the predictions of deep learning model by adding minimal changes to the examples. Only small changes are made to the data, so stable learning is possible.

To achieve a balance between the attack process and the maintenance of the semantic space is the main issue when utilizing the DATA scheme. In this research, DATA-based segmentation networks focused only on building extraction from high-resolution aerial images. If it is utilized with other datasets which contain multi-class objects such as roads, trees, ships, and other vehicles, to achieve the outcome of balanced adversarial attack while also maintaining the semantic space becomes more of a challenge. Integrating this issue into the learning phase is the focus of our upcoming work.

REFERENCES

[1] E. Maggiori, Y. Tarabalka, G. Charpiat, and P. Alliez, "Can semantic labeling methods generalize to any city? The Inria aerial image labeling benchmark," *Proc. IEEE Int. Geosci. Remote Sens. Symp.*, 2017, pp. 3226–3229.

[2] V. Mnih, "Machine learning for aerial image labeling," Ph.D. dissertation, Dept. Comput. Sci., Univ. Toronto, Toronto, ON, Canada, 2013.

[3] S. Ji, S. Wei, M. Lu, "Fully convolutional networks for multisource building extraction from an open aerial and satellite imagery data set", *IEEE Trans. Geosci. Remote Sens.*, vol. 57, no. 1, pp. 574-586, Jan. 2019.

[4] Guo, Yiyou, et al. "Global-Local Attention Network for Aerial Scene Classification." *IEEE Access* (2019).

[5] J. S. Blundell and D. W. Opitz, "Object recognition and feature extraction from imagery: The feature analyst approach," *Int. Arch. Photogram., Remote Sens. Spatial Inf. Sci.*, vol. 36, no. 4, p. C42, 2006.

[6] Y. LeCun, L. Bottou, Y. Bengio, and P. Haffner, "Gradient-based learning applied to document recognition," *Proc. IEEE*, vol. 86, no. 11, pp. 2278–2324, Nov. 1998.

[7] S. Chen, H. Wang, F. Xu, Y.-Q. Jin, "Target classification using the deep convolutional networks for SAR images", *IEEE Trans. Geosci. Remote Sens.*, vol. 54, no. 8, pp. 4806-4817, Jun. 2016.

[8] F. Luus, B. Salmon, F. Van Den Bergh, B. Maharaj, "Multiview deep learning for land-use classification", *IEEE Geosci. Remote Sens. Lett.*, vol. 12, no. 12, pp. 2448-2452, Dec. 2015.

[9] Y. Chen, X. Zhao, X. Jia, "Spectral-spatial classification of hyperspectral data based on deep belief network", *IEEE J. Sel. Topics Appl. Earth Observ. Remote Sens.*, vol. 8, no. 6, pp. 2381-2392, Jun. 2015.

[10] W. Li, H. Liu, Y. Wang, Z. Li, Y. Jia, G. Gui, "Deep learning-based classification methods for remote sensing images in urban built-up areas", *IEEE Access*, vol. 7, pp. 36274-36284, 2019.

[11] Y. Liu, Y. Zhong, Q. Qin, "Scene classification based on multiscale convolutional neural network", *IEEE Trans. Geosci. Remote Sens.*, vol. 56, no. 12, pp. 7109-7121, Dec. 2018.

[12] Q. Liu, R. Hang, H. Song, Z. Li, "Learning multiscale deep features for high-resolution satellite image scene classification", *IEEE Trans. Geosci. Remote Sens.*, vol. 56, no. 1, pp. 117-126, Jan. 2017.

[13] W. Diao, X. Sun, X. Zheng, F. Dou, H. Wang, K. Fu, "Efficient saliency-based object detection in remote sensing images using deep belief networks", *IEEE Geosci. Remote Sens. Lett.*, vol. 13, no. 2, pp. 137-141, Feb. 2016.

[14] X. Chen, S. Xiang, C.-L. Liu, C.-H. Pan, "Vehicle detection in satellite images by hybrid deep convolutional neural networks", *IEEE Geosci. Remote Sens. Lett.*, vol. 11, no. 10, pp. 1797-1801, Oct. 2014.

[15] Z. Lin, K. Ji, X. Leng, G. Kuang, "Squeeze and excitation rank faster R-CNN for ship detection in SAR images", *IEEE Geosci. Remote Sens. Lett.*, vol. 16, no. 5, pp. 751-755, May 2019.

[16] Z. Deng, H. Sun, S. Zhou, and J. Zhao, "Learning deep ship detector in SAR images from scratch", *IEEE Trans. Geosci. Remote Sens.*, vol. 57,no. 6, pp. 4021–4039, Jun. 2019.

[17] E. Maggiori, G. Charpiat, Y. Tarabalka, and P. Alliez, "Recurrent neural networks to correct satellite image classification maps," *IEEE Trans. Geosci. Remote Sens.*, vol. 55, no. 9, pp. 4962–4971, Sep. 2017.

[18] L. Mou, X. X. Zhu, RiFCN: Recurrent network in fully convolutional network for semantic segmentation of high resolution remote sensing images, 2018, [Online] Available: https://arxiv.org/abs/1805.02091.

[19] Li, Xiang, Xiaojing Yao, and Yi Fang. "Building-A-Nets: Robust Building Extraction from High-Resolution Remote Sensing Images with Adversarial Networks." *IEEE J. Sel. Topics Appl. Earth Observ. Remote Sens.*, vol. 11, no. 11, pp. 3680-3687, Aug. 2018.

[20] I. Goodfellow et al., "Generative adversarial nets", *Proc. Adv. Neural Inf. Process. Syst.*, pp. 2672-2680, 2014.

[21] Hu, Tao, et al. "SOBEL heuristic kernel for aerial semantic segmentation." *2018 25th IEEE International Conference on Image Processing (ICIP)*. IEEE, 2018.

[22] S. Gupta, S. G. Mazumdar, "Sobel edge detection algorithm", *Int. J. Comput. Sci. Manag. Res.*, vol. 2, no. 2, pp. 1578-1583, 2013.

[23] J. H. Kim, H. Lee, S. J. Hong, S. Kim, J. Park, J. Y. Hwang, and J. P. Choi, "Objects Segmentation From High-Resolution Aerial Images Using U-Net with Pyramid Pooling Layers." *IEEE Geosci. Remote Sens. Lett.*, vol. 16, no. 1, pp. 115–119, Jan. 2019.

[24] H. Zhao, J. Shi, X. Qi, X. Wang, J. Jia, "Pyramid scene parsing network", *Proc. IEEE Conf. Comput. Vis. Pattern Recognit. (CVPR)*, pp. 6230-6239, Jul. 2017.

[25] O. Ronneberger, P. Fischer, and T. Brox, "U-net: Convolutional networks for biomedical image segmentation," in *Proc. Med. Image Comput. Comput.-Assisted Intervention*, 2015, pp. 234–241.

[26] Y. Ganin, E. Ustinova, H. Ajakan, P. Germain, H. Larochelle, F. Laviolette, M. Marchand, V. S. Lempitsky, "Domain-adversarial training of neural networks", *J. Mach. Learn. Res.*, vol. 17, pp. 59:1-59:35, 2016.

[27] K. Bousmalis, N. Silberman, D. Dohan, D. Erhan, D. Krishnan, "Unsupervised pixel–level domain adaptation with generative adversarial networks", *Proc. CVPR*, pp. 3722-3731, 2017.

[28] Y.-H. Chen, W.-Y. Chen, Y.-T. Chen, B.-C. Tsai, Y.-C. F. Wang, M. Sun, "No more discrimination: Cross city adaptation of road scene segmenters", *Proc. Int. Conf. Comput. Vis.*, pp. 2011-2020, 2017.M.

[29] J. Hoffman, D. Wang, F. Yu, T. Darrell, FCNs in the wild: Pixel-level adversarial and constraint-based adaptation, 2016, [Online] Available: https://arxiv.org/abs/1612.02649.

[30] Y.-H. Tsai, W.-C. Hung, S. Schulter, K. Sohn, M.-H. Yang, M. Chandraker, "Learning to adapt structured output space for semantic segmentation", *Proc. IEEE Conf. Comput. Vis. Pattern Recognit.*, pp. 7472-7481, Jun. 2018.

[31] I. Goodfellow, NIPS 2016 tutorial: Generative adversarial networks, 2016, [Online] Available: https://arxiv.org/abs/1701.00160.

[32] S. Huang, N. Papernot, I. Goodfellow, Y. Duan, P. Abbeel, Adversarial attacks on neural network policies, 2017, [online] Available: https://arxiv.org/abs/1702.02284.

[33] I. J. Goodfellow, J. Shlens, C. Szegedy, Explaining and harnessing adversarial examples, 2014, [Online] Available: https://arxiv.org/abs/1412.6572.

[34] C. Szegedy et al., Intriguing properties of neural networks, 2013, [Online] Available: https://arxiv.org/abs/1312.6199.

[35] S. Baluja, I. Fischer, Adversarial transformation networks: Learning to generate adversarial examples, 2017, [Online] Available: https://arxiv.org/abs/1703.09387.

[36] R. Volpi, H. Namkoong, O. Sener, J. Duchi, V. Murino, S. Savarese. "Generalizing to unseen domains via adversarial data augmentation", *Proc. NIPS*, Dec. 2018.

[37] L. van der Maaten, G. Hinton, "Visualizing data using t-SNE", *J. Mach. Learn. Res.*, vol. 9, no. 11, pp. 2579-2605, 2008.

[38] J. Long, E. Shelhamer, T. Darrell, "Fully convolutional networks for semantic segmentation", *Proc. IEEE Conf. Comput. Vision Pattern Recog.*, 2015.

[39] T. M. Quan, D. G. Hildebrand, W.-K. Jeong, FusionNet: A deep fully residual convolutional neural network for image segmentation in connectomics, 2016, [Online] Available: https://arxiv.org/abs/1612.05360.

[40] J. Hu, L. Shen, G. Sun, Squeeze-and-excitation networks, 2017, [Online] Available: https://arxiv.org/abs/1709.01507.

[41] Y. Zhang, Y. Tian, Y. Kong, B. Zhong, Y. Fu, "Residual dense network for image super-resolution", *Proc. IEEE Conf. Comput. Vis. Pattern Recognit. (CVPR)*, pp. 2472-2481, Jun. 2018.

[42] G. Huang, Z. Liu, L. van der Maaten, K. Q. Weinberger, "Densely connected convolutional networks", *Proc. IEEE Conf. Comput. Vis. Pattern Recognit.*, vol. 1, pp. 2261-2269, 2017.

[43] Ioffe and C. Szegedy, "Batch normalization: Accelerating deep network training by reducing internal covariate shift," in *Proc. Int. Conf. Mach. Learn*, 2015, pp. 448–456.

[44] M. Abadi et al., "TensorFlow: A system for large-scale machine learning," in *Proc. OSDI*, vol. 16, 2016, pp. 265–283.

[45] D. P. Kingma, J. Ba, Adam: A method for stochastic optimization., 2014, [Online] Available: https://arxiv.org/abs/1412.6980.

[46] M. A. Rahman, Y. Wang, "Optimizing intersection-over-union in deep neural networks for image segmentation", *Proc. 12th Int. Symp. I Adv. Vis. Comput. (ISVC)*, pp. 234-244, Dec. 2016.

[47] Lu, Kangkang, Ying Sun, and Sim-Heng Ong. "Dual-Resolution U-Net: Building Extraction from Aerial Images." *2018 24th International Conference on Pattern Recognition (ICPR)*. IEEE, 2018.

[48] Inria Aerial Image Labeling Benchmark Leader Board. [Online] Available: https://project.inria.fr/aerialimagelabeling/leaderboard/.

[49] B. Chatterjee and C. Poullis, "On Building Classification from Remote Sensor Imagery Using Deep Neural Networks and the Relation Between Classification and Reconstruction Accuracy Using Border Localization as Proxy", *Proc. Conference on Computer and Robot Vision (CRV)*, pp. 41-48, 2019.